\documentclass{article}
\usepackage{arxiv}

\usepackage[utf8]{inputenc} % allow utf-8 input
\usepackage[T1]{fontenc}    % use 8-bit T1 fonts
\usepackage{hyperref}       % hyperlinks
\usepackage{url}            % simple URL typesetting
\usepackage{booktabs}       % professional-quality tables
\usepackage{amsfonts}       % blackboard math symbols
\usepackage{nicefrac}       % compact symbols for 1/2, etc.
\usepackage{microtype}      % microtypography
\usepackage{lipsum}

\usepackage[toc,page]{appendix}
\usepackage[ruled,vlined]{algorithm2e}
\usepackage{multicol}
\usepackage{multirow}
\usepackage{graphicx}
\usepackage{array}
\usepackage{float}
\usepackage{bm}
\usepackage{biblatex}
\DeclareNameAlias{author}{family-given}
\usepackage[labelfont=bf]{caption}

\title{A Memory-Augmented Neural Network Model of Abstract Rule Learning}

\author{\hspace{-0.95cm}
  Ishan Sinha\\\hspace{-0.95cm}
  Department of Computer Science\\\hspace{-0.95cm}
  Princeton University\\\hspace{-0.95cm}
  \texttt{isinha@alumni.princeton.edu}\\
  \And
\hspace{-0.95cm}  
Taylor W. Webb\\\hspace{-0.95cm}
  Department of Psychology\\\hspace{-0.95cm}
  University of California, Los Angeles\\\hspace{-0.95cm}
  \texttt{taylor.w.webb@gmail.com}\\\hspace{-0.95cm}
  \And
\hspace{-0.95cm}
  Jonathan D. Cohen\\\hspace{-0.95cm}
  Department of Neuroscience\\\hspace{-0.95cm}
  Princeton University\\\hspace{-0.95cm}
  \texttt{jdc@princeton.edu}\\\hspace{-0.95cm}
}

\begin{document}

\maketitle

\begin{abstract} \normalsize
Human intelligence is characterized by a remarkable ability to infer abstract rules from experience and apply these rules to novel domains. As such, designing neural network algorithms with this capacity is an important step toward the development of deep learning systems with more human-like intelligence. However, doing so is a major outstanding challenge, one that some argue will require neural networks to use explicit symbol-processing mechanisms. In this work, we focus on neural networks' capacity for arbitrary role-filler binding, the ability to associate abstract “roles” to context-specific “fillers,” which many have argued is an important mechanism underlying the ability to learn and apply rules abstractly. Using a simplified version of Raven’s Progressive Matrices, a hallmark test of human intelligence, we introduce a sequential formulation of a visual problem-solving task that requires this form of binding. Further, we introduce the Emergent Symbol Binding Network (ESBN), a recurrent neural network model that learns to use an external memory as a binding mechanism. This mechanism enables symbol-like variable representations to emerge through the ESBN's training process without the need for explicit symbol-processing machinery. We empirically demonstrate that the ESBN successfully learns the underlying abstract rule structure of our task and perfectly generalizes this rule structure to novel fillers.
\end{abstract}

\section{Introduction}
Humans exhibit a unique ability to adapt and apply knowledge and learned rules to unfamiliar domains. For example, a basketball coach may apply his knowledge about spacing on the playing field to create a game plan for a soccer team, even though the specific playing fields in the two sports vastly differ. Similarly, an expert chess player may apply general strategies from chess to play a game of Go, despite the significant differences between the parameters of these games. In essence, abstract rule learning involves learning general strategies from experiences in a manner that allows application of those concepts to new contexts.

It has long been argued that the ability to apply knowledge to new domains is a hallmark of human intelligence \cite{gentner1983structure, marcus2001algebraic, lake2017building}. It follows, then, that for deep learning systems to exhibit human-like intelligence, they must possess this capacity. However, while neural networks excel - and often even surpass humans - at tasks requiring application of rules to familiar domains, a growing literature indicates that current systems are largely limited to this form of generalization and fail to apply rules to new contexts \cite{LakeBaroni, BarrettPGM, hill2019learning}.

In this work, we study, and attempt to improve, the ability of neural networks to learn and apply rules abstractly. To do so, we turn to Raven's Progressive Matrices (RPM), a task that is commonly used to assess abstract reasoning capacity. Introduced in 1938, RPM is among the most widely used tests of so-called “fluid” intelligence, the ability to reason about and solve novel problems \cite{RPM, Snow, Carpenter}. In RPM (Figure 1 below), test takers are presented with a $3\times3$ matrix in which each of the first two rows contains a distinct set of images with common abstract relationships. For example, the size of shapes may decrease along the rows. The final image of the third row is omitted, and based on the abstract relationships at play in the first two rows, the participant must choose the best image from eight candidate images to fill in the blank. While humans are able to complete many RPM and RPM-like problems after little or no exposure, recent work shows that neural networks struggle significantly with these problems even after a substantial degree of training, especially when required to generalize to problems that involve components on which they have not been trained \cite{BarrettPGM, hill2019learning, Zhang_2019, zhang2019learning, hu2020hierarchical}.

To remedy this disparity, we propose a technique that enables neural networks to carry out \textit{arbitrary role-filler binding}. Role-filler binding involves representation in terms of abstract “roles” and context-specific fillers that “fill” these roles. For example, in the sentence, "I am a human," the subject of the sentence (role) is "I" (filler). It has been argued that this representational strategy is a critical component of the human capacity for abstract reasoning, and is also a key property by which traditional symbolic approaches achieve their generality \cite{Holyoak, indirection, hummel_holyoak_1, hummel_holyoak_2, hummel_holyoak_3}. In particular, the ability to \textit{arbitrarily} bind fillers to abstract roles enables a powerful form of generalization. Armed with this capacity, an agent can, in principle, learn a task in a manner that is abstracted away from the particular fillers that figure in its training examples, inferring the underlying abstract rule in a way such that it can apply that rule to entirely novel fillers. 

To test for a neural network’s ability to perform arbitrary role-filler binding, we created a dataset implementing an RPM-like, visual problem solving task that required the application of a learned, abstract rule to fillers not observed in the training set (Figure 2 below). The dataset contained a range of generalization conditions, varying the number of fillers that were withheld during training but maintained in testing. 

To approach the development of a model with this capacity, we turn to neural networks augmented with external memory. We propose a model, the Emergent Symbol Binding Network (ESBN), in which memories consist of explicit bindings between roles and fillers, but in which the content of the roles and fillers, and the way in which they are used, are learned from experience. Specifically, the model employs two segregated information processing streams -- one in which a recurrent controller generates vectors that ultimately act as roles, and one that stores learned embeddings of the fillers in our dataset. These two streams only interact through bindings in the external memory, enabling the controller to solve the task in a manner that is agnostic of the specific fillers. As a result, this binding mechanism allows symbol-like representations to emerge through the training process, equipping the model to learn abstract problem structure without requiring built-in symbol-processing mechanisms. We evaluated this model on our visual problem solving task and benchmarked it against a standard Long Short-Term Memory network (LSTM) \cite{lstm} and a simplified Neural Turing Machine (NTM) \cite{NTM}. We found that, unlike the LSTM and NTM, the ESBN displayed near perfect generalization to novel fillers, even in the most extreme generalization conditions of our dataset.

\section{Related Work}

\subsection{Raven's Progressive Matrices and Deep Learning}

There have been a number of attempts at using deep learning to solve RPM. Barrett et al. \cite{BarrettPGM} introduced the Procedurally-Generated Matrices (PGM) dataset, a large dataset of automatically generated RPM-like matrices, and a novel variant of the Relation Net \cite{relationnet} to attempt the task. Zhang et al. \cite{Zhang_2019} published RAVEN, a large dataset of automatically generated matrices that more closely follows the original RPM rules than the PGM dataset, as well as a neural module to improve a model’s structured reasoning capability. Zhang et al. \cite{zhang2019learning} used a contrast effect in training and a Noise-Contrastive Estimation loss function in their Contrastive Perceptual Inference network, which they tested on PGM and RAVEN. Hill et al. \cite{hill2019learning} took a somewhat different approach by focusing on the way in which a model is trained, introducing the LABC (Learning Analogies by Contrasting Abstract Relational Structure) training method. Hu et al. \cite{hu2020hierarchical} introduced the Balanced-RAVEN dataset to correct biases in Zhang et al.’s RAVEN dataset \cite{Zhang_2019}, and they introduced the Hierarchical Rule Induction Network. Wu et al. \cite{wu2020scattering} introduced the Scattering Compositional Learner (SCL), an architecture that attempts to capitalize on the compositional structure of RPM problems, and tested it on the PGM and Balanced-RAVEN datasets. Finally, Jahrens and Martinetz \cite{jahrens2020solving} combined Barrett et al.'s Wild Relation Network with Multi-Layer Relation Networks \cite{jahrens2018multilayer}, introduced a novel encoding scheme, and tested the architecture on the PGM dataset.

These studies have made impressive progress on the basic challenge of designing models that are capable of solving RPM-like problems. In particular, the SCL \cite{wu2020scattering} achieved 95.0\% accuracy on Balanced-RAVEN, and Martinetz's Multi-Layer Relation Networks \cite{jahrens2020solving} achieved 98\% accuracy on the neutral regime of the PGM dataset. However, these models typically require very large training sets (on the order of $10^{6}$ training examples), and generally fail to generalize outside of the very specific conditions under which they are trained. For example, state-of-the-art performance on the \textit{extrapolation} regime of the PGM dataset, in which test problems contain feature values outside the range of those observed in the training set, is currently 25.9\% \cite{wang2020generalisable}, and state-of-the-art performance on other out-of-distribution generalization regimes (\textit{held-out shape-color}, \textit{held-out line-type}, etc.) is comparably poor \cite{BarrettPGM, wang2020abstract}. This stands in sharp contrast to human learners, who are often able to solve RPM problems zero-shot.

Our work differs in two ways. First, in our dataset, we used simplified RPM-like problems, stripping away unnecessary complexity in order to focus on the specific feature of generalization to novel fillers (analogous to Barrett et al.’s \textit{held-out line-type} and \textit{held-out shape-color} regimes \cite{BarrettPGM}). Doing so allowed us to focus our modeling efforts on this problem. Second, unlike other models, our model showed near perfect generalization to novel fillers. 

There are also a number of symbolic approaches that have been proposed for solving RPM problems, including an influential model from Carpenter et al. \cite{Carpenter} and a recent extension of the Structure Mapping Engine \cite{falkenhainer, lovettforbus}. Our goal is somewhat different than these approaches -- to learn abstract rules and symbol-like representations from experience, rather than building this capacity in by hand.

\subsection{Memory-Augmented Neural Networks}

There are various proposals for augmenting neural networks with an external memory. An influential early line of work, Complementary Learning Systems, proposes that neural systems benefit from having components that learn on different time scales, and argues that this combination allows neural networks both to learn general, abstract structure, and to rapidly encode new items \cite{complementarylearningsystems}. A recent technique, Fast Weights, proposes an efficient, end-to-end method for doing this, by combining a recurrent neural network with a fast learning auto-associative memory \cite{ba2016using}. Another recently proposed architecture, the Tolman-Eichenbaum Machine, uses a technique similar to Fast Weights to model various aspects of the entorhinal and hippocampal system \cite{Tolman}. In a related, but simplified and abstracted, approach found in models such as the NTM, memories are stored as entries in an external memory matrix that a controller network reads from and writes to, much like how a Turing Machine interacts with a tape \cite{NTM}. It has been shown that this sort of external memory architecture can effectively generalize rules to longer sequences than those observed during training \cite{NTM} and can facilitate one-shot learning \cite{santoro_oneshot_2016}. A similar approach, the Differentiable Neural Dictionary (DND), also utilizes an external memory matrix, but separates memories into distinct "keys" and "values" \cite{DND}. 

Conceptually, our approach is similar to that of the Tolman-Eichenbaum Machine, in that we also represent memories in a manner that separates abstract structure from concrete fillers, and do so by employing segregated information processing pathways. However, the mechanisms we employ are more closely related to the DND, though the content of the items in memory, the exact mechanisms for interacting with the memory, and the goal of the model, are quite different. Relative to this previous work, our central contribution is to provide a method that infers an abstract rule directly from a sequence of high-dimensional data (images), and does so in a manner that generalizes to entirely novel fillers. Chen et al. \cite{Chen} showed that their implementation of an NTM was capable of a similar form of generalization. However, the fillers in their study were randomly-generated, continuous 50-dimensional vectors, such that, given a sufficiently dense sampling of the space, the fillers presented during testing might have been similar enough to fillers observed during training to permit a form of interpolation. To avoid this possibility, as described in Section 3.3 below, we used higher-dimensional fillers ($32 \times 32$ images) and sometimes sampled only 3 or 5 fillers from the whole space for training, resulting in a greater degree of disjointness between fillers in the training and test sets. Nevertheless, we included the NTM as a benchmark model to test its generalization capability in our task.

\section{Task}

\begin{figure}[htbb]
\begin{minipage}[b]{0.475\linewidth}
\centering
\includegraphics[width=.75\linewidth]{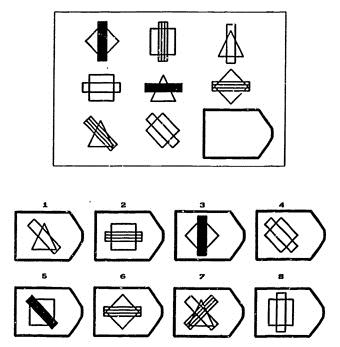}
\caption{Raven's Progressive Matrices problem \normalfont{\cite{Carpenter}}. {\normalfont This example employs two instances of the “distribution-of-three“ rule. The correct answer is option 5.}}
\label{fig:side-a}
\end{minipage}
\hspace{0.5cm}
\begin{minipage}[b]{0.475\linewidth}
\centering
\includegraphics[width=.75\linewidth]{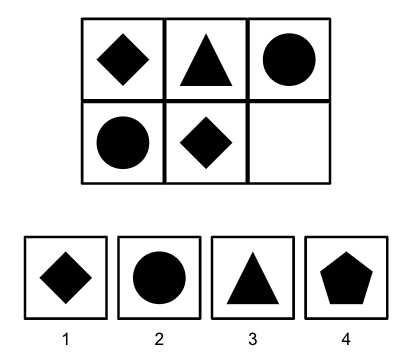}
\caption{Simplified Raven's Progressive Matrices problem. {\normalfont This example employs a single instance of the “distribution-of-three“ rule. The correct answer is option 3.}}
\label{fig:side-b}
\end{minipage}
\vspace{0.15in}
\label{figure1}
\end{figure}

\subsection{Visual Problem Solving Task}

We created a visual problem solving task inspired by Raven's Progressive Matrices. Our task utilized the "distribution-of-three" rule as characterized by Carpenter et al. \cite{Carpenter}. This rule employs a set-theoretic relation which requires that “three values from a categorical attribute (such as figure type) are distributed through a row” \cite{Carpenter}, but without regard for the order in which these values appear. For example, in the problem in Figure 2, a diamond, triangle, and circle are present in the first row. According to the rule, these same shapes must be present in the second row, so the missing cell should contain a triangle. In this example, the roles can be thought of as "Shape 1," "Shape 2," and "Shape 3," and their respective fillers are diamond, triangle, and circle. 

In our task, a model was given the 5 non-empty cells in a matrix and had to either generate the filler that belonged in the blank cell or select that filler from the four multiple choice (MC) options. For each problem, the four MC options consisted of the three unique fillers from the first row (one of which was the correct answer) and another randomly selected filler, and they were arranged in random order.

We formulated the problem as a sequential task involving two alternative task modes. In both modes, the three cells in the first row and first two cells in the second row were presented to a model in sequence. In the \textit{generative} mode, following the presentation of this sequence, the task was to generate the correct answer. In the example in Figure 2, a model would need to generate a triangle to complete the matrix. Alternatively, in \textit{multiple choice (MC)} mode, a model was additionally presented the four multiple choice options in sequence and required to output the index of the correct option. In the example in Figure 2, a model would need to output index three. We investigated both of these task modes because human test takers use both strategies to solve RPM problems \cite{BethellFox, Carpenter} and because the generative mode is a stronger test of generalization performance than MC mode.

\subsection{Generation of Examples}

We first created sequences of integers in which each integer represented a filler in a cell of the matrix, and we then uniquely assigned each integer to either a one-hot vector or a unicode image. Specifically, to generate trials for $N$ fillers, we first generated all sets of three integers from the set, $\{0,\ 1,\ \ldots,\ N - 1\}$, of which there are $N \choose 3$. Then, for each combination, we generated all \textit{3!} of its permutations. Each of these permutations was one possible ordering of the fillers in a row of the matrix. Since there are two rows, there were \textit{3!3!}=36 possible matrix problems per each set of three integers. We chose not to create multiple examples that only differed by the randomly selected fourth MC option. Thus, there were $36$ $N\choose3$ total possible examples with $N$ fillers. From this pool, we randomly sampled without replacement the desired number of examples. Finally, we assigned each integer either to a unique one-hot vector or to a unique unicode image.

\subsection{Data Types}

\subsubsection{One-Hot Vectors}
First, we experimented with \textit{N-}dimensional one-hot vectors (where $N = 100$) because they are simple and genuinely distinct due to their mutual orthogonality. In the generative mode, a model received a sequence of $5$ \textit{N-}dimensional one-hot vectors as input and outputted the target \textit{N-}dimensional vector. In MC mode, a model received a sequence of $9$ \textit{N-}dimensional one-hot vectors as input and outputted the target MC answer as a four-dimensional categorical distribution over the four MC options.  

\subsubsection{Grayscale Unicode Images}
We then selected $N = 100$ visually distinct, grayscale, $32 \times 32$ unicode images (see Appendix C for all images) to test whether different models might have different generalization properties with respect to fully orthogonal stimuli versus ones with graded similarity, and also to assess whether our proposed approach could be extended to higher-dimensional data. In the generative mode, a model received a sequence of five $32\times32$ images as input and outputted a $32\times32$ grayscale image. We then calculated the predicted image’s nearest neighbor over the set of 100 images by using mean squared error as a distance metric, and used that nearest neighbor image to calculate accuracy. In MC mode, a model received a sequence of nine $32\times32$ images as input and outputted the target MC answer as a four-dimensional categorical distribution over the four MC options.

\subsection{Generalization Regime}

Our generalization regime tested models exclusively on fillers to which they had not been exposed in training. Successful generalization to novel fillers, we argue, is an exceptionally strong - and perhaps the strongest - measure of a model’s capacity for arbitrary role-filler binding. 

We used $N = 100$ fillers and withheld $M \in \{0,\ 50,\ 85,\ 95,\ 97\}$ of these fillers from training to be used exclusively for testing. When $M = 0$, we generated both training and testing examples with the integer set $\{0,\ 1,\ldots,\ N - 1\}$ through the process outlined in Section 3.2, but we ensured that the training and test sets were disjoint (i.e., even though the two sets involved the same fillers, those fillers were arranged into distinct sequences). For $M \neq 0$, we generated training examples with the integer set $\{0,\ 1,\ldots,\ N - M - 1\}$ and testing examples with the integer set $\{N - M,\ N - M + 1,\ldots,\ N - 1\}$. We include the final sizes of the train and test sets that we sampled in Appendix A.

When $M = 0$, a model was tested on novel permutations of fillers that it had already seen in training. This regime was independently and identically distributed insofar as each of the $N$ fillers had a roughly equal probability of appearing in any given training or testing example. To successfully generalize in this condition, a model needed to learn the underlying distribution-of-three rule and have the capacity for basic role-filler binding since it had not encountered any of the exact testing problems before. We are hesitant to describe this as \textit{arbitrary} role-filler binding, however, since at some point in the course of training, the model had seen each filler and even had seen most fillers in each cell of the visual $2\times3$ matrix. 

For $M > 3$, a model was tested exclusively on fillers that it had not seen in training, requiring genuine abstract rule application to novel fillers. To successfully generalize, a model needed to learn the underlying rule and use true arbitrary role-filler binding since it needed to represent the roles in the problem in an abstract enough way to bind them to unfamiliar fillers. By progressively withholding a larger proportion of fillers from training, we tested a model's ability to learn the rule in an abstract manner. The ultimate test was the $M=97$ case, in which a model was trained on the minimum number of fillers for the problem to be possible - a single set of three. We could not attempt the task in MC mode with $M=97$ because that mode required at least four fillers to serve as MC options.

\section{Models}

\subsection{Autoencoder}

\subsubsection{Motivation}
The generalization regimes in which $M > 3$ might seem infeasible with one-hot vectors as fillers since the weights corresponding to the withheld fillers in a network’s input layer would not be trained. This was one motivation for using images: since they all lie in $32\times32$ space, all of a network’s weights would likely be trained even when novel images were used in testing. To address this issue in a manner that allowed us to employ one-hots, we trained an autoencoder over the entire space of fillers. For consistency and completeness, we also used an autoencoder for images and for all values of $M$. Thus, each model had an \textit{autoencoding} component that handled dimensionality conversion for inputs and outputs, and a \textit{sequential} component (i.e., the LSTM, NTM, or ESBN) that learned the inferential procedure to accomplish the task. The autoencoder can be thought of as a "front-end" that processed the raw input for the sequential component and converted the sequential component's output to the desired filler space. Specifically, the encoding layers of the autoencoder component were applied to all inputs, and the decoding layers were used for generative prediction.

\subsubsection{Integration with the Sequential Component of the Model}
Let $f_s$ denote the sequential component of the model, $f_e$ and $f_d$ denote the encoder and decoder modules of the autoencoder component, respectively, and $\boldsymbol{x}_{t=1}$\ldots$\boldsymbol{x}_{t=T}$ denote the input sequence. 

At each time step $t$, a model could use the decoder, $f_d$, to make a prediction about the input at the next time step, $\hat{\boldsymbol{x}}_{t+1}$. In the generative mode, we presented a model with a sequence consisting of the $T=5$ non-empty cells in a particular matrix problem, and trained the model to generate a prediction for the final cell:

\[\hat{\boldsymbol{x}}_{t=T+1} = f_d (f_s (f_e(\boldsymbol{x}_{t=1}),\ \ldots\ ,f_e(\boldsymbol{x}_{t=T})))\]

Each model also had the capacity to generate task-specific outputs, $\hat{y_t}$, at each time step. In MC mode, we presented a model with a sequence of length $T=9$, consisting of the $5$ non-empty cells in a particular matrix problem and the corresponding $4$ MC options, and trained the system to select the correct MC option by using the task output at the last time step:

\[\hat{\boldsymbol{y}}_{t=T} = f_s (f_e(\boldsymbol{x}_{t=1}),\ \ldots\ ,f_e(\boldsymbol{x}_{t=T}))\]

\subsubsection{Vector Autoencoder}
The autoencoder for one-hot vectors was a simple feedforward network. It encoded an \textit{N-}dimensional input ($N = 100$ in our experiments), $\boldsymbol{x}$, into a 10-dimensional embedding, \boldsymbol{$z$}, by passing it through a Fully-Connected (FC) ReLU layer. It decoded an embedding, $\boldsymbol{z}$, by passing it through a FC softmax layer with $N$ units.
\\
\subsubsection{Image Autoencoder}
The autoencoder for unicode images encoded a $32 \times\ 32$ input image, $\boldsymbol{x}$, by passing it through three convolutional layers, each with 32 kernels of size $4\times4$ and a stride of 2 (no max-pooling), resulting in a feature map of size $4\times4\times32$. These were followed by 2 FC layers with 256 and 128 units, respectively, resulting in a 128-dimensional embedding, $\boldsymbol{z}$. It decoded an embedding by first passing it through two FC layers with 256 and 512 units, respectively. The output of these FC layers was reshaped into a feature map of size $4\times4\times32$ and then passed through three transposed convolutional layers, each with 32 kernels of size $4\times4$ and a fractional stride of 1/2 (no max-pooling). Each layer of the autoencoder had a ReLU nonlinearity besides the final decoder layer, which used a Sigmoid nonlinearity so that the decoded image was grayscale. In some cases (detailed in Appendix A), batch normalization \cite{ioffe2015batch} was applied before the nonlinearity of each layer in order to accelerate training.

\subsection{Emergent Symbol Binding Network}

\subsubsection{Controller}
The Emergent Symbol Binding Network (ESBN; Algorithm~\ref{ESBN_algo}) uses an LSTM controller with a differentiable external memory that is explicitly separated into keys and values, corresponding to roles and fillers. The keys are generated by an output layer from the LSTM controller, and the values are the individual input embeddings, $\boldsymbol{z}_t$, of the input sequence. The model learns how to represent roles in the keys it generates and how to bind those roles to fillers according to its training objective. The model also learns key and value "gates," or scalar parameters that control to what extent it retrieves from memory during each time step of the input sequence (described below). These gates are soft, ranging between 0 and 1, so that the whole system is fully differentiable. Our controller implementation had 1 recurrent layer and 512 features in its hidden state.

As input for time step $t$, the controller receives $\boldsymbol{h}_{t-1}$, its initial hidden state and cell state for the time step, and $\boldsymbol{k}_{r_{t-1}}$, a Read Key vector that is derived from the keys in the external memory at time $t-1$ (described below). The controller outputs:
\begin{enumerate}
    \item $g_{k_t}$, $g_{v_t}$ -- Key and Value Gate scalars -- through a FC sigmoidal layer
    \item $\boldsymbol{k}_{q_t}$ -- a Query Key vector that is used to retrieve values from memory -- through a FC ReLU layer
    \item $\boldsymbol{k}_{w_t}$ -- a Write Key vector that is written to memory alongside $\boldsymbol{z}_t$ -- through a FC ReLU layer 
\end{enumerate} 
In our implementation, $\boldsymbol{k}_{r}$, $\boldsymbol{k}_q$ and $\boldsymbol{k}_{w}$ were 256-dimensional. The schematic in Figure 3 illustrates these outputs and the model at large. 

\subsubsection{Reading}
The model uses indirection, the use of one entity to refer to another (similar to a pointer in low-level programming), to retrieve keys and values from memory. Almost all computer systems use indirection to achieve variable binding, and it is argued to be a neurally-plausible mechanism by which variable binding is implemented in the brain as well \cite{indirection}.

The specific procedure mirrors that of the content-focused reading mechanism of the NTM \cite{NTM} and DND \cite{DND}, as well as the soft attention mechanisms that are now ubiquitous in NLP sequence models \cite{vaswani}. In each time step $t$, a measure of similarity (we used the dot product) is calculated between the Query Key, $\boldsymbol{k}_{q_t}$, and all keys in memory, $\boldsymbol{M}_{k_{t-1}}$, and these outputs are scaled to the $(0,1)$-range with a softmax. The scaled dot products are then used as weights to take a weighted average of the \textit{values} in memory, $\boldsymbol{M}_{v_{t-1}}$. Unlike the NTM and DND, this weighted average value is scaled by the value gate parameter, $g_{v_t}$, resulting in a latent space prediction for the input at the next time step, $\hat{\boldsymbol{z}}_{t+1}$. The gate parameter effectively allows the model to choose \textit{when} to retrieve from memory.

Analogously, the Read Key vector, $\boldsymbol{k}_{r_t}$, is derived from $\boldsymbol{M}_{k_{t-1}}$ in each time step by using the similarity between $\boldsymbol{z}_t$ and $\boldsymbol{M}_{v_{t-1}}$. 

\subsubsection{Writing}
At the end of each time step $t$, the key ($\boldsymbol{k}_{w_t}$) and value ($\boldsymbol{z}_t$) are simply concatenated row-wise to the existing key memories ($\boldsymbol{M}_{k_{t-1}}$) and value memories ($\boldsymbol{M}_{v_{t-1}}$), respectively. Therefore, if $\boldsymbol{k}_{w_t}$ and $\boldsymbol{z}_t$ are $d$-dimensional and $f$-dimensional, respectively, then 
$\boldsymbol{M}_{k_t}$ is of size $t \times\ d$ and $\boldsymbol{M}_{v_t}$ is of size $t \times\ f$.

Technically, $\boldsymbol{M}_k$ and $\boldsymbol{M}_v$ do have infinite capacity, though they have at most 9 entries in the task modes we present. This is a proxy for the limitless external store which is generally thought to exist in the brain \cite{Beukers_limitless}. In this model, much like in the theories of the hippocampus and episodic memory that inspire it \cite{complementarylearningsystems, Tolman}, the primary functional constraint is on retrieval, not storage.

\subsubsection{Algorithm}
\begin{algorithm}[H]
\SetAlgoLined
 $\boldsymbol{k}_{r_{t=0}} \leftarrow\ \boldsymbol{0}$\;
 $\boldsymbol{h}_{t=0} \leftarrow\ \boldsymbol{0}$\;
 $\boldsymbol{M}_{k_{t=0}} \leftarrow\ \{\}$\;
 $\boldsymbol{M}_{v_{t=0}} \leftarrow\ \{\}$\;
\For{\normalfont{$t$ in $1$ \ldots $T$}}{
    $\boldsymbol{z}_t \leftarrow\ f_e(\boldsymbol{x}_t)$\;
    $\hat{\boldsymbol{y}}_{t}, \
    g_{k_t}, \
    g_{v_t}, \
    \boldsymbol{k}_{w_t}, \
    \boldsymbol{k}_{q_t}, \
    \boldsymbol{h}_{t} \
    \leftarrow\  f_s(\boldsymbol{h}_{t-1}, \ \boldsymbol{k}_{r_{t-1}})$\;
    \eIf{$t$ is $1$}{
        $\boldsymbol{k}_{r_{t}} \leftarrow\ \boldsymbol{0}$\;
        $\hat{\boldsymbol{z}}_{t+1} \leftarrow\ \boldsymbol{0}$\;
    }{
        $\boldsymbol{w}_{k_{t}} \leftarrow\ $ softmax($\boldsymbol{M}_{v_{t-1}} \cdot \
        \boldsymbol{z}_{t}$) \;
        $\boldsymbol{k}_{r_{t}} \leftarrow\ g_{k_t} \displaystyle\sum_{i=1}^{t-1} w_{k_{t}}(i) \boldsymbol{M}_{k_{t-1}}(i)$ \;
        $\boldsymbol{w}_{v_{t}} \leftarrow\ $ softmax($\boldsymbol{M}_{k_{t-1}} \cdot \
        \boldsymbol{k}_{q_{t}}$) \;
        $\hat{\boldsymbol{z}}_{t+1} \leftarrow\ g_{v_t} \displaystyle\sum_{i=1}^{t-1} w_{v_t}(i) \boldsymbol{M}_{v_{t-1}}(i)$\;
    }
    $\boldsymbol{M}_{k_{t}} \leftarrow\ \{\boldsymbol{M}_{k_{t-1}}, \ \boldsymbol{k}_{w_{t}}\}$\;
    $\boldsymbol{M}_{v_{t}} \leftarrow\ \{\boldsymbol{M}_{v_{t-1}}, \ \boldsymbol{z}_t\}$\;
    $\hat{\boldsymbol{x}}_{t+1} \leftarrow\  f_d(\hat{\boldsymbol{z}}_{t+1}$)\;
 }
 \textbf{return} $\hat{\boldsymbol{y}}_{t=T}$, $\hat{\boldsymbol{x}}_{t=T+1}$
 \label{ESBN_algo}
 \caption{Emergent Symbol Binding Network. As in Section 4.1.2, let $f_s$ denote the sequential component of the model, $f_e$ and $f_d$ denote the encoder and decoder modules of the autoencoder component, respectively, $\boldsymbol{x}_{t=1}$\ldots$\boldsymbol{x}_{t=T}$ denote the input sequence, and $\hat{\boldsymbol{x}}_{t}$ and $\hat{\boldsymbol{y}}_{t}$ denote generative and MC predictions, respectively. $\{ , \}$ denotes the concatenation of a matrix and a vector, resulting in a matrix with one additional row.}
\end{algorithm}
\vspace{0.1 in}

\noindent We allowed the ESBN to run for an extra time step (resulting in $T=6$ for the generative mode and $T=10$ for MC mode) to allow it to process the Read Key, $\boldsymbol{k}_{r_{T-1}}$, that was retrieved from memory when viewing the final embedding of the input sequence, $\boldsymbol{z}_{T-1}$. During this extra time step, no additional input embedding was presented to the network (so $f_e$ was not executed), and no key was retrieved from memory, since this would not have affected the model's final response.

\begin{figure}[h!]
\centering
\includegraphics[width=\linewidth]{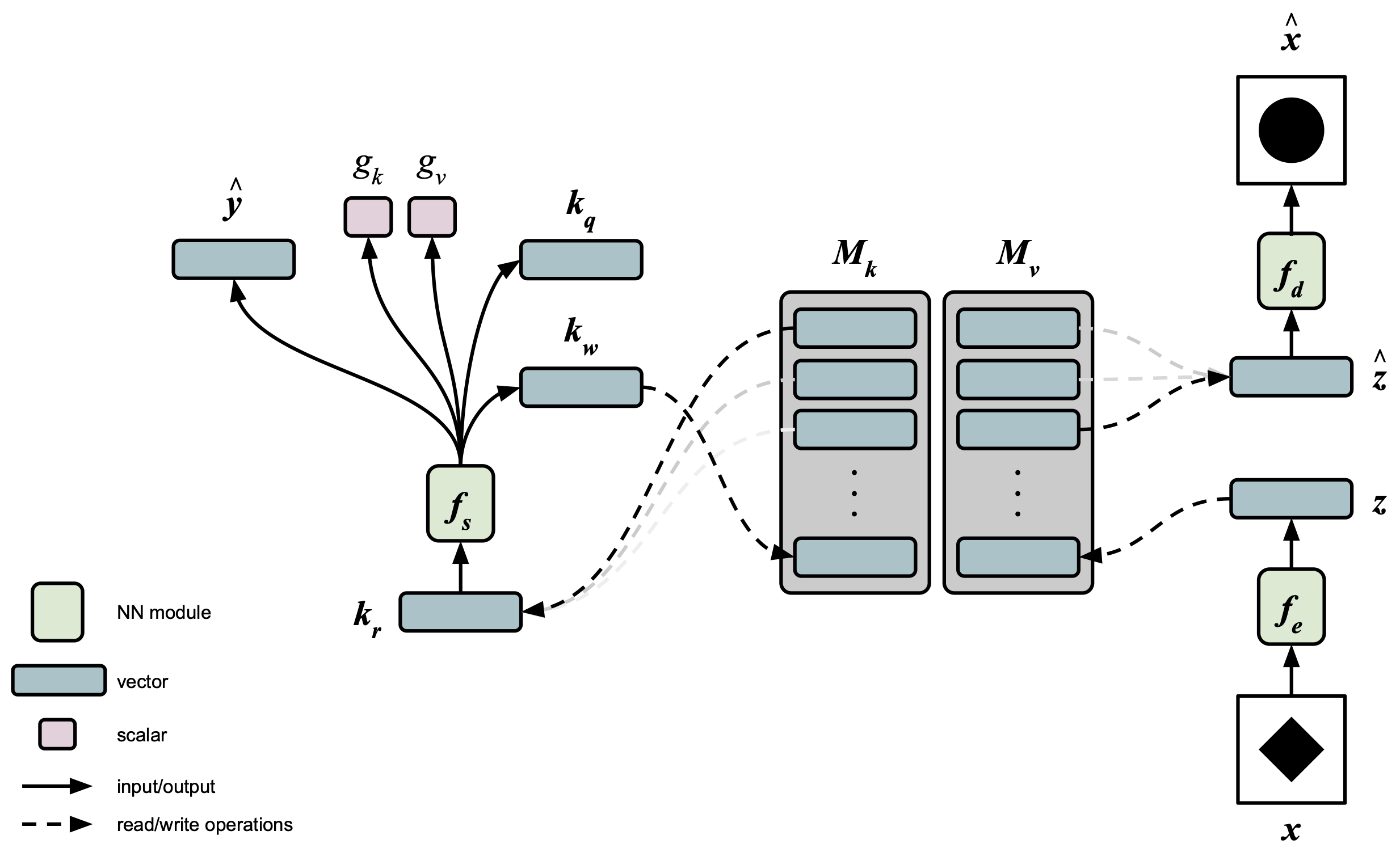}
\caption{Schematic of the Emergent Symbol Binding Network. {\normalfont The Query Key, $\boldsymbol{k}_{q}$, and the Value Gate, $g_{v}$, are used to derive a latent space prediction, $\hat{\boldsymbol{z}}$, from $\boldsymbol{M}_{v}$. Similarly, the encoded input, $\boldsymbol{z}$, and the Key Gate, $g_{k}$, are used to derive the Read Key, $\boldsymbol{k}_{r}$, from $\boldsymbol{M}_{k}$. The Write Key, $\boldsymbol{k}_{w}$, and $\boldsymbol{z}$ are concatenated to $\boldsymbol{M}_{k}$ and $\boldsymbol{M}_{v}$, respectively. The model can make a generative prediction, $\hat{\boldsymbol{x}}$, through the Decoder, $f_{d}$, or a MC prediction, $\hat{\boldsymbol{y}}$, through a task output layer from the LSTM controller. These predictions are trained with separate loss functions.}}
\label{figure2}
\end{figure}

\subsection{Simplified Neural Turing Machine}
We benchmarked our model against the NTM, an influential external memory architecture, in part because Chen et al. \cite{Chen} showed it has some potential to generalize to novel fillers. Unlike the ESBN, the NTM has one fixed-size memory matrix instead of a dynamically-growing memory that is separated into keys and values. In Appendix A, we detail how we initialized this memory matrix. Due to the fixed memory size, the NTM also has a mechanism for deciding \textit{where} to write. 

Our implementation of the NTM used an LSTM controller with 1 recurrent layer and 512 features in its hidden state. After experimentation, we used a memory size of $10 \times\ 256$, 256-dimensional Add, Erase, Write Key and Read Key vectors, and a ReLU nonlinearity for the Add, Erase, Write Key and Read Key vectors. We implemented the content-focused reading mechanism as described in Graves et al. \cite{NTM}, except that we used the dot product instead of cosine similarity as our similarity measure, as we did in the ESBN. We omitted the location-based reading mechanism, implemented the writing and erase mechanisms as described, and used one write head and one read head.

\subsection{Standard LSTM}
We also benchmarked against a standard LSTM \cite{lstm}. Our implementation had 1 recurrent layer and 512 features in its hidden state.

\subsection{Pre-trained System vs. End-to-End System}

Our central aim in this work is to study generalization to novel fillers, but fillers can be defined as novel in one of two senses: 1) novel in the context of a specific task vs. 2) completely novel, in the sense that the agent has never experienced them in any context. An example of the former is someone using a hammer to split lumber when he had previously only used hammers on nails. An example of the latter is someone using a hammer to split lumber without having ever used one before. We operationalized these two scenarios by either 1) pre-training the autoencoder or 2) training it end-to-end with the main model. 

In the former method, we pre-trained the autoencoder over the entire space of $N$ fillers, froze its parameters, and then trained the sequential component of the model. This way, the model had never seen the withheld fillers in the context of the visual problem solving task, but it had seen them before the context of this task. We believe this may be more closely aligned with how the brain works, insofar as a person may have had visual experience with (e.g., retinal processing of) stimuli for which he has never had to carry out a problem-solving task requiring the form of arbitrary binding that is necessary for the task performed by our model (and that is thought to be implemented in the hippocampus). 

In the latter method, we instantiated the autoencoder but did not train it or freeze its weights, and then trained the sequential and autoencoding components of the model together. This method tested generalization to data that was completely outside the domain of the training data on which all modules of the system had been trained. As a result, this required a significantly stronger form of generalization than when the autoencoder was pre-trained. In particular, generating predictions for novel fillers in testing was especially difficult because the decoder never learned how to construct these fillers.

Henceforth, we use the nomenclature \textit{pre-trained} to refer to a model with a pre-trained, frozen autoencoder and \textit{end-to-end} to refer to a model in which both components learned together.

\subsection{Training Parameters}

For all models and generalization regimes, we conducted a hyperparameter search for the learning rate, number of training epochs, and whether or not to apply batch normalization before each nonlinearity in the autoencoder. To evaluate all models as fairly as possible, we searched the same hyperparameters for each model and selected the best hyperparameters for each model according to training accuracy. In all but one case, models were more than 90\% accurate on the training set, typically 99-100\% accurate. We detail the final training hyperparameters and our initialization schemes for the models in Appendix A, and we detail the training performance results in Appendix B.

\section{Experimental Results}

In this section, we report test accuracy and associated standard error over 10 trained networks with different random seeds. We detail the results further in Appendix B.

\subsection{Results for Pre-trained Models}

\subsubsection{One-Hot Vectors}

\begin{figure}[htbb]
\begin{minipage}[b]{0.475\linewidth}
\centering
\includegraphics[width=\linewidth]{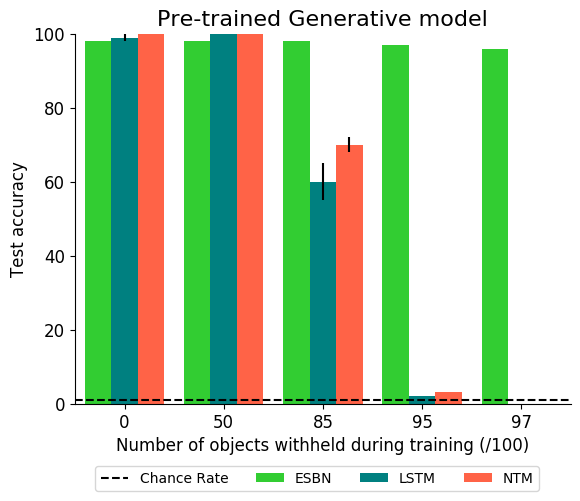}
\caption{Generalization Accuracy for Pre-trained, Generative Models on One-Hot Vectors}
\label{fig2:side-a}
\end{minipage}
\hspace{0.5cm}
\begin{minipage}[b]{0.475\linewidth}
\centering
\includegraphics[width=\linewidth]{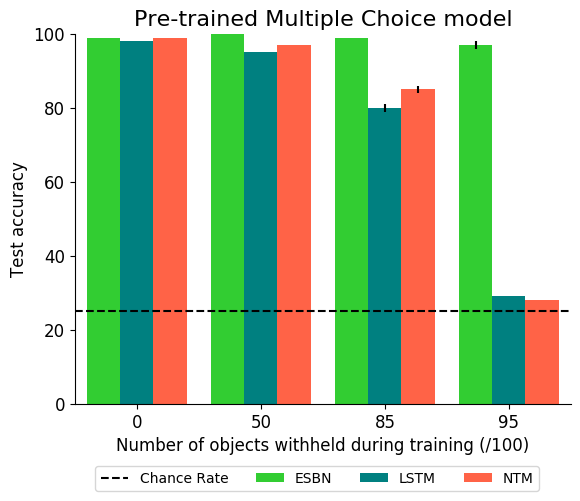}
\caption{Generalization Accuracy for Pre-trained, MC Models on One-Hot Vectors}
\label{fig2:side-b}
\end{minipage}
\label{figure3}
\end{figure}

Figures 4 and 5 depict the generalization accuracy for the three models for different values of $M$. These results show that both for the generative and MC variants of the task, the ESBN maintained near perfect generalization accuracy despite the number of withheld fillers, indicating that it solved the task through true arbitrary role-filler binding. Remarkably, even in the $M=97$ case in which there were only 3 fillers and 36 examples in training, it still generalized nearly perfectly. This indicates that the model can extract an abstract rule from limited experience, as humans often do. 

In both modes, the LSTM and NTM achieved near perfect generalization accuracy when no fillers were withheld from training ($M = 0$), suggesting that these models are capable of learning basic role-filler binding. However, test accuracy monotonically decreased as the number of withheld shapes increased, eventually reaching chance performance when 95 out of 100 shapes were held out. Since generalization accuracy did not immediately fall to chance rate when fillers were withheld, these results do indicate that the LSTM and NTM have some capacity for abstract rule learning and quasi-arbitrary role-filler binding. However, the gradual degradation of generalization accuracy suggests that these models ultimately required filler-specific representations to accomplish the objective.

The NTM did show minor gains over the LSTM in generalization accuracy for $M \in \{50,\ 85\}$, which indicates that it did use its external memory in a way that allowed it to learn the problem marginally more abstractly. In the Discussion section of this paper, we explore why the NTM did not generalize as well as the ESBN. 
\\
\subsubsection{Unicode Images}

\begin{figure}[htbb]
\begin{minipage}[b]{0.475\linewidth}
\centering
\includegraphics[width=\linewidth]{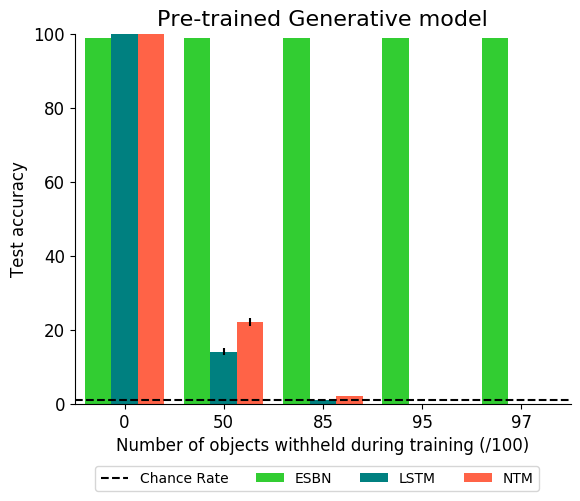}
\caption{Generalization Accuracy for Pre-trained, Generative Models on Unicode Images}
\label{fig3:side-a}
\end{minipage}
\hspace{0.5cm}
\begin{minipage}[b]{0.475\linewidth}
\centering
\includegraphics[width=\linewidth]{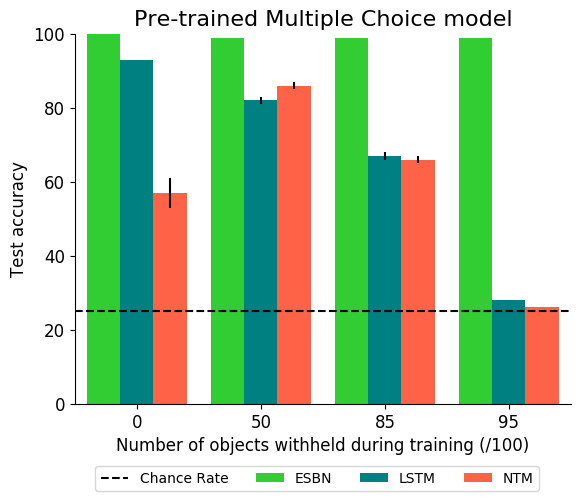}
\caption{Generalization Accuracy for Pre-trained, MC Models on Unicode Images}
\label{fig3:side-b}
\end{minipage}
\label{figure4}
\end{figure}

\begin{figure}[htp]
\centering
\includegraphics[width=.3\textwidth]{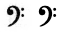}\hfill
\includegraphics[width=.3\textwidth]{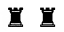}\hfill
\includegraphics[width=.3\textwidth]{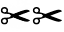}
\caption{Randomly Selected Examples of Generated (Left) vs. Target (Right) Images}
\vspace{0.1in}
\label{figure5}
\end{figure}

Figures 6 and 7 show that the ESBN again had near perfect generalization accuracy despite the number of fillers withheld from training. Figure 8 shows randomly selected examples of accurate, generated images, side-by-side with their respective target images. The differences are almost indecipherable. We provide more examples of generated images in Appendix D.

There are two differences with the LSTM and NTM results here compared to those with the one-hot vectors. First, the test accuracy with $M \in \{50,\ 85\}$ was much lower. We hypothesize that this occurred because the images covered a larger, more complex space than the one-hot vectors. Second, the NTM was unable to generalize when no fillers were withheld in MC mode, and even had better accuracy with $M \in \{50,\ 85\}$. This is especially puzzling, given that it had relative success in learning this task when it was trained end-to-end, as described below. We include this as a point for further research.

\subsection{Results for End-to-End Models}

We did not train models end-to-end with one-hot vectors because the weights corresponding to the withheld fillers in a model's encoder module would not have been trained. 

\begin{figure}[htbb]
\begin{minipage}[b]{0.475\linewidth}
\centering
\includegraphics[width=\linewidth]{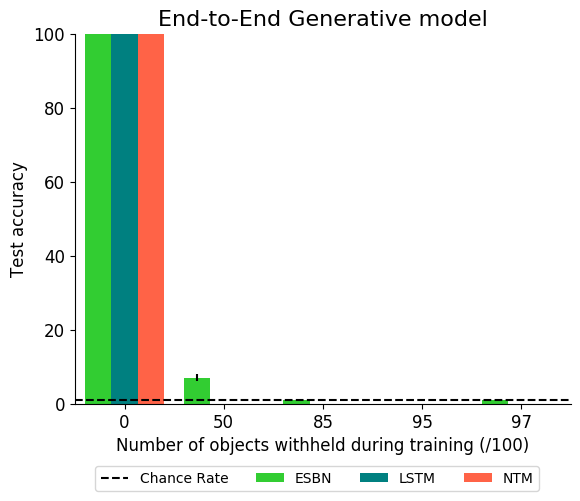}
\caption{Generalization Accuracy for End-to-End, Generative Models on Unicode Images}
\label{fig4:side-a}
\end{minipage}
\hspace{0.5cm}
\begin{minipage}[b]{0.475\linewidth}
\centering
\includegraphics[width=\linewidth]{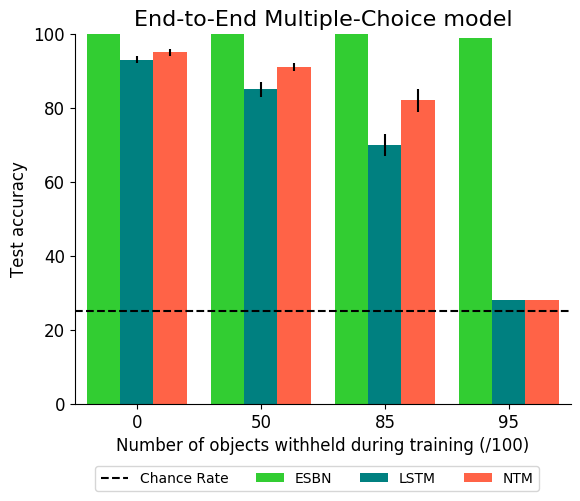}
\caption{Generalization Accuracy for End-to-End, MC Models on Unicode Images}
\label{fig4:side-b}
\end{minipage}
\label{figure6}
\end{figure}

In the generative mode (Figure 9), all models were able to perform basic role-filler binding to generalize when $M = 0$. None of the models were able to accomplish the task when images were withheld, presumably because the decoder never learned how to construct the withheld images. We do note that the ESBN displayed marginally above-chance accuracy in the $M=50$ case.

In MC mode (Figure 10), the ESBN again showed near perfect generalization accuracy across the board despite the fact that no part of the model had been exposed to the test set fillers in any context. These fillers were entirely novel. This is the most compelling evidence we present, and it strongly suggests the model learns the problem through genuine arbitrary role-filler binding. 

In MC mode, the LSTM and NTM's generalization accuracies were high for $M = 0$ but decreased as $M$ increased. This again suggests that the LSTM and NTM are capable of basic role-filler binding, but that they did not use true arbitrary role-filler binding to solve the task. Interestingly, the LSTM and NTM were able to generalize here much better than in the pre-trained versions of the models (Figure 7). Since training accuracy was high in both versions (see Appendix B), this indicates that the end-to-end model has the unique advantage of learning latent representations that are more suited to the specific constraints of our task. 

\section{Discussion and Further Research}

In this work, we introduce a dataset with a simple visual problem-solving task that tests a neural network’s capacity for arbitrary role-filler binding. We find that the standard LSTM and the NTM are incapable of the level of abstract reasoning that is necessary to solve the task with unseen fillers. We introduce the Emergent Symbol Binding Network, and we demonstrate that it learns how to accomplish the task with true arbitrary role-filler binding such that it can successfully generalize to novel fillers. The most remarkable result is that it was able to accomplish the multiple choice variant of the task on unseen fillers even when trained end-to-end. Moreover, it was able to do so while only being trained on 360 examples with only 5 unique fillers, approaching the sample efficiency of human learners (though we note that humans are capable of solving RPM problems in some cases with no prior exposure at all). This finding, along with the model's success in both the pre-trained and end-to-end training conditions, displays the robustness of the approach. 

Our findings indicate that it is important to implement indirection, by separating memory into two distinct components and using one to refer to the other, to achieve arbitrary role-filler binding. Both the ESBN and NTM models used the same content-focused mechanism to read from memory, so the defining difference likely was the structure of the external memory. Namely, the explicit segregation of information processing streams allowed the Emergent Symbol Binding Network to generalize perfectly out of domain, whereas the NTM, with its single memory matrix, only had marginal gains over the LSTM. 

While these results indicate that this memory factorization is an important inductive prior, the exact implementation is yet to be crystallized. For example, further research may determine whether it can be introduced in a softer manner than it is in our current approach, and whether it is an architectural constraint on external memory that should generally be imposed.

We hope that our findings here promote and influence further research towards development of models that can reason abstractly. In particular, our results indicate that future attempts at problems like Raven's Progressive Matrices should incorporate external memory mechanisms. We hope to bolster this claim by extending our research to other datasets and tasks. For example, the MNIST and ImageNet datasets afford the opportunity to train and test models on different instances of each image class, whereas our unicode image dataset had only one instance per class. This would challenge models to learn the distribution-of-three rule in a manner that is abstract with respect to both the task and the filler space. We also are interested in extending our work to other RPM relations, such as distribution-of-two and pairwise progression \cite{Carpenter}, and existing RPM datasets \cite{BarrettPGM, Zhang_2019, hu2020hierarchical}. In a similar vein, we are also interested to see how some of the proposed models that we reference would perform on our dataset.

There is a gray area with our research with regard to categorizing a filler as truly "novel" since distinctness is often defined on a spectrum (e.g., "somewhat distinct" versus "very distinct"), especially in image space. However, we think that the mutual orthogonality of one-hot vectors and the obvious visual distinctness of the images we selected (along with the fact that we only chose 100 images out of the enormous space of possible images) reasonably substantiate the novelty of the withheld fillers in our tasks. Extensions of our research to datasets like those mentioned above should further test this claim.

The model we present also has implications for theories of Working Memory (WM) and Episodic Memory (EM). Some might think of our system as a model of WM due to the short-term and activity-based nature of the bindings involved in the task. However, our model is not consistent with a sole WM interpretation because the bindings are not dynamic -- once they have been created, they cannot be erased and re-written. WM undoubtedly plays a role in our model since the recurrent processing in an LSTM is typically interpreted in terms of WM, but the ESBN's variable binding mechanisms, as well as those of many other external memory architectures, are based on neuroscientific theories about EM and the hippocampus' role in generalization. Therefore, the data and model we present are consistent with a growing literature that indicates that many tasks that have been traditionally thought to only involve WM also incorporate EM \cite{Beukers_WM_has_EM}. 

It is also worth considering how the forms of variable binding we see in this task may occur in the brain in absence of rows and tensor concatenation. The first major proposal is that there is an establishment of a semi-permanent synaptic connection in the hippocampus that physically links two different neural populations \cite{CerOReilley}. Another proposal, that leans more towards working memory, is that neurons in different areas of the brain form a binding while firing in synchrony but lose this binding while not in synchrony \cite{CerOReilley}. Since our model does not have a mechanism for dynamically unbinding, it seems to be more consistent with the former hypothesis. However, further work will be needed to determine how exactly binding is implemented in the brain. Our work here demonstrates that the inner workings of such binding mechanism may help explain humans' remarkable capacity for rapid generalization of rules.

We look forward to contributing further research that helps neural networks achieve human-like rule learning and hope that our work, in conjunction with others’, can shed light on the mechanisms that enable abstract reasoning.

%%%%%%%%%%%%%%%%
% \bibliographystyle{unsrt}  
% \bibliography{references}
\printbibliography
%%%%%%%%%%%%%%%%

\clearpage
\begin{appendices}

\section{Experimental Setup Details}

\subsection{Model Initialization}

We initialized all layers according to their activation functions as follows:
\begin{itemize}
	\item Tanh: Xavier Normal \cite{Glorot2010UnderstandingTD} with $gain=5/3$
	\item Sigmoid: Xavier Normal with $gain=1$
	\item ReLU: Kaiming Normal \cite{Kaiming} with $gain=sqrt(2)$
	\item Softmax: Xavier Normal with $gain=1$
\end{itemize}
\vspace{0.1in}
According to this framework, we initialized the input-to-hidden and hidden-to-hidden weights separately from the gate weights in the LSTM and LSTM controllers.
\\\\
We used Xavier Normal initialization for the the NTM's memory matrix and treated it as a learned parameter to be optimized over training. At the beginning of the forward pass for each training batch, we set the memory matrix equal to the initialized memory matrix. This way, the initialization changed over the course of training but was the same for each batch in testing. 

\subsection{Examples per Regime}

\begin{table}[hbt]
  \centering
  \begin{tabular}{|c|c|c|} \hline
    \textbf{M} & \textbf{Train Examples} & \textbf{Test Examples}\\\hline
    0 & 10000 & 10000\\\hline
    50 & 10000 & 10000\\\hline
    85 & 10000 & 10000\\\hline
    95 & 360 & 10000\\\hline
    97 & 36 & 10000\\\hline
  \end{tabular}
  \caption{Number of Examples with $N = 100$}
  \label{table:data}
\end{table}

\subsection{Training Parameters}

For each model and regime, we conducted a hyperparameter search for the learning rate, number of training epochs, and whether or not to apply batch normalization before each nonlinearity in the autoencoder for unicode images. For fairness, we searched the same hyperparameters for each model, and we selected the best for each model according to training accuracy.

Each entry in Tables 2-7 below consists of the learning rate, number of training epochs, and whether or not the autoencoder used batch normalization (T or F). 

We typically used a significantly greater number of epochs for $M \in \{95,\ 97\}$ because an epoch for this condition involved far fewer training iterations than for smaller values of $M$ (see Table 1).
\\
\subsubsection{One-Hot Vectors}\hspace*{\fill} \\
We used the ADAM optimizer \cite{kingma2014adam} and Cross Entropy Loss across the board. Each model used a batch size of 32.
\\\\
\noindent \underline{Autoencoder (Pre-trained models only)}: We used a learning rate of 0.01 for 500 epochs with a batch size of 10.
\\
\begin{table}[!ht]  
  \centering
  \begin{tabular}{c|c|c|c|c|c}
    \cline{2-6}
    \multicolumn{1}{c|}{} & \textbf{M = 0} & \textbf{M = 50} & \textbf{M = 85} & \textbf{M = 95} & \textbf{M = 97}\\\hline
    \textbf{LSTM} & $5e^-5$, 80, F & $5e^-5$, 80, F & $5e^-5$, 80, F & $5e^-4$, 1500, F & $5e^-4$, 2000, F\\\hline
    \textbf{NTM} & $5e^-5$, 80, F & $5e^-5$, 80, F & $5e^-5$, 80, F & $5e^-5$, 1500, F & $5e^-5$, 2000, F\\\hline
    \textbf{ESBN} & $5e^-5$, 80, F & $5e^-5$, 80, F & $5e^-5$, 80, F & $5e^-5$, 1500, F & $5e^-5$, 2000, F\\\hline
  \end{tabular}
  \caption{Training Hyperparameters for Pre-trained, Generative Models on One-Hot Vectors.}
\end{table}

\begin{table}[!ht]  
  \centering
  \begin{tabular}{c|c|c|c|c}
    \cline{2-5}
    \multicolumn{1}{c|}{} & \textbf{M = 0} & \textbf{M = 50} & \textbf{M = 85} & \textbf{M = 95}\\\hline
    \textbf{LSTM} & $5e^-4$, 50, F & $5e^-4$, 50, F & $5e^-4$, 50, F & $5e^-4$, 50, F \\\hline
    \textbf{NTM} & $5e^-4$, 50, F & $5e^-4$, 50, F & $5e^-4$, 50, F & $5e^-4$, 50, F \\\hline
    \textbf{ESBN} & $5e^-4$, 50, F & $5e^-4$, 50, F & $5e^-4$, 50, F & $5e^-4$, 50, F \\\hline
  \end{tabular}
  \caption{Training Hyperparameters for Pre-trained, MC Models on One-Hot Vectors.}
\end{table}

\subsubsection{Unicode Images}\hspace*{\fill} \\
We used the ADAM optimizer across the board, Mean Squared Error Loss between the predicted and target images for generative prediction, and Cross Entropy Loss for MC prediction. Each model used a batch size of 32.
\\\\
\noindent \underline{Autoencoder (Pre-trained models only)}: We used a learning rate of $5e^-4$ for 500 epochs with a batch size of 10.

\begin{table}[H]  
  \centering
  \begin{tabular}{c|c|c|c|c|c}
    \cline{2-6}
    \multicolumn{1}{c|}{} & \textbf{M = 0} & \textbf{M = 50} & \textbf{M = 85} & \textbf{M = 95} & \textbf{M = 97}\\\hline
    \textbf{LSTM} & $5e^-5$, 200, F & $5e^-5$, 200, F & $5e^-5$, 200, F & $5e^-5$, 1500, F & $5e^-5$, 2500, F\\\hline
    \textbf{NTM} & $5e^-5$, 150, F & $5e^-5$, 150, F & $5e^-5$, 150, F & $5e^-5$, 1500, F & $5e^-5$, 2500, F\\\hline
    \textbf{ESBN} & $5e^-5$, 50, F & $5e^-5$, 50, F & $5e^-5$, 50, F & $5e^-5$, 1500, F & $5e^-5$, 2500, F\\\hline
  \end{tabular}
  \caption{Training Hyperparameters for Pre-trained, Generative Models on Unicode Images.}
\end{table}

\begin{table}[H]  
  \centering
  \begin{tabular}{c|c|c|c|c}
    \cline{2-5}
    \multicolumn{1}{c|}{} & \textbf{M = 0} & \textbf{M = 50} & \textbf{M = 85} & \textbf{M = 95}\\\hline
    \textbf{LSTM} & $5e^-4$, 200, T & $5e^-4$, 200, T & $5e^-4$, 200, T & $5e^-4$, 1500, T\\\hline
    \textbf{NTM} & $5e^-4$, 150, T & $5e^-4$, 150, T & $5e^-4$, 150, T & $5e^-4$, 1500, T\\\hline
    \textbf{ESBN} & $5e^-5$, 50, F & $5e^-5$, 50, F & $5e^-5$, 50, F & $5e^-5$, 1500, F\\\hline
  \end{tabular}
  \caption{Training Hyperparameters for Pre-trained, MC Models on Unicode Images.}
\end{table}

\begin{table}[H]  
  \centering
  \begin{tabular}{c|c|c|c|c|c}
    \cline{2-6}
    \multicolumn{1}{c|}{} & \textbf{M = 0} & \textbf{M = 50} & \textbf{M = 85} & \textbf{M = 95} & \textbf{M = 97}\\\hline
    \textbf{LSTM} & $5e^-4$, 100, F & $5e^-4$, 150, F & $5e^-4$, 100, F & $5e^-4$, 100, F & $5e^-4$, 150, F\\\hline
    \textbf{NTM} & $5e^-4$, 100, F & $5e^-4$, 100, F & $5e^-4$, 100, F & $5e^-4$, 100, F & $5e^-4$, 150, F\\\hline
    \textbf{ESBN} & $5e^-5$, 250, T & $5e^-5$, 250, T & $5e^-5$, 250, T & $5e^-5$, 1500, T & $5e^-5$, 2500, T\\\hline
  \end{tabular}
  \caption{Training Hyperparameters for End-to-End, Generative Models on Unicode Images.}
\end{table}

\begin{table}[H]  
  \centering
  \begin{tabular}{c|c|c|c|c}
    \cline{2-5}
    \multicolumn{1}{c|}{} & \textbf{M = 0} & \textbf{M = 50} & \textbf{M = 85} & \textbf{M = 95}\\\hline
    \textbf{LSTM} & $5e^-4$, 150, F & $5e^-4$, 100, F & $5e^-4$, 100, F & $5e^-4$, 100, F\\\hline
    \textbf{NTM} & $5e^-4$, 150, F & $5e^-4$, 150, F & $5e^-4$, 100, F & $5e^-4$, 100, F\\\hline
    \textbf{ESBN} & $5e^-5$, 150, T & $5e^-5$, 150, T & $5e^-5$, 150, T & $5e^-5$, 1500, T\\\hline
  \end{tabular}
  \caption{Training Hyperparameters for End-to-End, MC Models on Unicode Images.}
\end{table}

\clearpage
\section{Detailed Experimental Results}
Each entry in the following tables consists of the average percentage accuracy and associated standard error (to two significant figures) over 10 trained networks with different random seeds. 

\subsection{Pre-trained Models}
\subsubsection{One-Hot Vectors}\hspace*{\fill}
\vspace{-0.15in}
\begin{table}[H]
  \centering
  \begin{tabular}{c|c|c|c|c|c|c}
    \cline{3-7}
     \multicolumn{1}{c}{} & \multicolumn{1}{c|}{} & \textbf{M = 0} & \textbf{M = 50} & \textbf{M = 85} & \textbf{M = 95} & \textbf{M = 97}\\\hline
     
    \multirow{2}*{\textbf{LSTM}} & Train Accuracy & 100 $\pm$ 0 & 100 $\pm$ 0 & 100 $\pm$ 0 & 100 $\pm$ 0 & 77 $\pm$ 7\\\cline{2-7}
    & Test Accuracy & 99 $\pm$ 1 & 100 $\pm$ 0 & 60 $\pm$ 5 & 2 $\pm$ 0 & 0 $\pm$ 0\\\hline
    
    \multirow{2}*{\textbf{NTM}} & Train Accuracy & 100 $\pm$ 0 & 100 $\pm$ 0 & 100 $\pm$ 0 & 98 $\pm$ 2 & 93 $\pm$ 4\\\cline{2-7}
    & Test Accuracy & 100 $\pm$ 0 & 100 $\pm$ 0 & 70 $\pm$ 2 & 3 $\pm$ 0 & 0 $\pm$ 0\\\hline
    
    \multirow{2}*{\textbf{ESBN}} & Train Accuracy & 99 $\pm$ 0 & 99 $\pm$ 0 & 99 $\pm$ 0 & 100 $\pm$ 0 & 100 $\pm$ 0\\\cline{2-7}
    & Test Accuracy & 98 $\pm$ 0 & 98 $\pm$ 0 & 98 $\pm$ 0 & 97 $\pm$ 0 & 96 $\pm$ 0\\\hline
    
  \end{tabular}
  \caption{Percentage Accuracies for Pre-trained, Generative Models on One-Hot Vectors.}
\end{table}

\vspace{-0.15in}
\begin{table}[H]
  \centering
  \begin{tabular}{c|c|c|c|c|c}
  \cline{3-6}
     \multicolumn{1}{c}{} & \multicolumn{1}{c|}{} & \textbf{M = 0} & \textbf{M = 50} & \textbf{M = 85} & \textbf{M = 95}\\\hline
     
    \multirow{2}*{\textbf{LSTM}} & Train Accuracy & 100 $\pm$ 0 & 100 $\pm$ 0 & 100 $\pm$ 0 & 100 $\pm$ 0\\\cline{2-6}
    & Test Accuracy & 98 $\pm$ 0 & 95 $\pm$ 0 & 80 $\pm$ 1 & 29 $\pm$ 0\\\hline
    
    \multirow{2}*{\textbf{NTM}} & Train Accuracy & 100 $\pm$ 0 & 100 $\pm$ 0 & 100 $\pm$ 0 & 100 $\pm$ 0\\\cline{2-6}
    & Test Accuracy & 99 $\pm$ 0 & 97 $\pm$ 0 & 85 $\pm$ 1 & 28 $\pm$ 0\\\hline
    
    \multirow{2}*{\textbf{ESBN}} & Train Accuracy & 100 $\pm$ 0 & 100 $\pm$ 0 & 100 $\pm$ 0 & 100 $\pm$ 0\\\cline{2-6}
    & Test Accuracy & 99 $\pm$ 0 & 100 $\pm$ 0 & 99 $\pm$ 1 & 97 $\pm$ 1\\\hline
    
  \end{tabular}
  \caption{Percentage Accuracies for Pre-trained, MC Models on One-Hot Vectors.}
\end{table}

\subsubsection{Unicode Images}\hspace*{\fill}

\vspace{-0.15in}
\begin{table}[H]
  \centering
  \begin{tabular}{c|c|c|c|c|c|c}
    \cline{3-7}
     \multicolumn{1}{c}{} & \multicolumn{1}{c|}{} & \textbf{M = 0} & \textbf{M = 50} & \textbf{M = 85} & \textbf{M = 95} & \textbf{M = 97}\\\hline
     
    \multirow{2}*{\textbf{LSTM}} & Train Accuracy & 100 $\pm$ 0 & 100 $\pm$ 0 & 100 $\pm$ 0 & 100 $\pm$ 0 & 100 $\pm$ 0\\\cline{2-7}
    & Test Accuracy & 100 $\pm$ 0 & 14 $\pm$ 1 & 1 $\pm$ 0 & 0 $\pm$ 0 & 0 $\pm$ 0\\\hline
    
    \multirow{2}*{\textbf{NTM}} & Train Accuracy & 100 $\pm$ 0 & 100 $\pm$ 0 & 100 $\pm$ 0 & 100 $\pm$ 0 & 100 $\pm$ 0\\\cline{2-7}
    & Test Accuracy & 100 $\pm$ 0 & 22 $\pm$ 1 & 2 $\pm$ 0 & 0 $\pm$ 0 & 0 $\pm$ 0\\\hline
    
    \multirow{2}*{\textbf{ESBN}} & Train Accuracy & 99 $\pm$ 0 & 100 $\pm$ 0 & 100 $\pm$ 0 & 100 $\pm$ 0 & 100 $\pm$ 0\\\cline{2-7}
    & Test Accuracy & 99 $\pm$ 0 & 99 $\pm$ 0 & 99 $\pm$ 0 & 99 $\pm$ 0 & 99 $\pm$ 0\\\hline
    
  \end{tabular}
  \caption{Percentage Accuracies for Pre-trained, Generative Models on Unicode Images.}
\end{table}

\vspace{-0.15in}
\begin{table}[H]
  \centering
  \begin{tabular}{c|c|c|c|c|c}
  \cline{3-6}
     \multicolumn{1}{c}{} & \multicolumn{1}{c|}{} & \textbf{M = 0} & \textbf{M = 50} & \textbf{M = 85} & \textbf{M = 95}\\\hline
     
    \multirow{2}*{\textbf{LSTM}} & Train Accuracy & 100 $\pm$ 0 & 100 $\pm$ 0 & 100 $\pm$ 0 & 100 $\pm$ 0\\\cline{2-6}
    & Test Accuracy & 93 $\pm$ 0 & 82 $\pm$ 1 & 67 $\pm$ 1 & 28 $\pm$ 0\\\hline
    
    \multirow{2}*{\textbf{NTM}} & Train Accuracy & 93 $\pm$ 7 & 100 $\pm$ 0 & 100 $\pm$ 0 & 91 $\pm$ 4\\\cline{2-6}
    & Test Accuracy & 57 $\pm$ 4 & 86 $\pm$ 1 & 66 $\pm$ 1 & 26 $\pm$ 0\\\hline
    
    \multirow{2}*{\textbf{ESBN}} & Train Accuracy & 100 $\pm$ 0 & 100 $\pm$ 0 & 100 $\pm$ 0 & 100 $\pm$ 0\\\cline{2-6}
    & Test Accuracy & 100 $\pm$ 0 & 99 $\pm$ 0 & 99 $\pm$ 0 & 99 $\pm$ 0\\\hline
    
  \end{tabular}
  \caption{Percentage Accuracies for Pre-trained, MC Models on Unicode Images.}
\end{table}

\subsection{End-to-End Models}

\begin{table}[H]
  \centering
  \begin{tabular}{c|c|c|c|c|c|c}
    \cline{3-7}
     \multicolumn{1}{c}{} & \multicolumn{1}{c|}{} & \textbf{M = 0} & \textbf{M = 50} & \textbf{M = 85} & \textbf{M = 95} & \textbf{M = 97}\\\hline
     
    \multirow{2}*{\textbf{LSTM}} & Train Accuracy & 100 $\pm$ 0 & 99 $\pm$ 1 & 100 $\pm$ 0 & 100 $\pm$ 0 & 100 $\pm$ 0\\\cline{2-7}
    & Test Accuracy & 100 $\pm$ 0 & 0 $\pm$ 0 & 0 $\pm$ 0 & 0 $\pm$ 0 & 0 $\pm$ 0\\\hline
    
    \multirow{2}*{\textbf{NTM}} & Train Accuracy & 100 $\pm$ 0 & 99 $\pm$ 1 & 100 $\pm$ 0 & 100 $\pm$ 0 & 100 $\pm$ 0\\\cline{2-7}
    & Test Accuracy & 100 $\pm$ 0 & 0 $\pm$ 0 & 0 $\pm$ 0 & 0 $\pm$ 0 & 0 $\pm$ 0\\\hline
    
    \multirow{2}*{\textbf{ESBN}} & Train Accuracy & 100 $\pm$ 0 & 100 $\pm$ 0 & 100 $\pm$ 0 & 100 $\pm$ 0 & 100 $\pm$ 0\\\cline{2-7}
    & Test Accuracy & 100 $\pm$ 0 & 7 $\pm$ 1 & 1 $\pm$ 0 & 0 $\pm$ 0 & 1 $\pm$ 0\\\hline
  \end{tabular}
  \caption{Percentage Accuracies for End-to-End, Generative Models on Unicode Images.}
\end{table}

\vspace{-0.15in}
\begin{table}[H]
  \centering
  \begin{tabular}{c|c|c|c|c|c}
  \cline{3-6}
     \multicolumn{1}{c}{} & \multicolumn{1}{c|}{} & \textbf{M = 0} & \textbf{M = 50} & \textbf{M = 85} & \textbf{M = 95}\\\hline
     
    \multirow{2}*{\textbf{LSTM}} & Train Accuracy & 99 $\pm$ 0 & 100 $\pm$ 0 & 100 $\pm$ 0 & 100 $\pm$ 0\\\cline{2-6}
    & Test Accuracy & 93 $\pm$ 1 & 85 $\pm$ 2 & 70 $\pm$ 3 & 28 $\pm$ 0\\\hline
    
    \multirow{2}*{\textbf{NTM}} & Train Accuracy & 98 $\pm$ 0 & 99 $\pm$ 0 & 100 $\pm$ 0 & 100 $\pm$ 0\\\cline{2-6}
    & Test Accuracy & 95 $\pm$ 1 & 91 $\pm$ 1 & 82 $\pm$ 3 & 28 $\pm$ 0\\\hline
    
    \multirow{2}*{\textbf{ESBN}} & Train Accuracy & 100 $\pm$ 0 & 100 $\pm$ 0 & 100 $\pm$ 0 & 100 $\pm$ 0\\\cline{2-6}
    & Test Accuracy & 100 $\pm$ 0 & 100 $\pm$ 0 & 100 $\pm$ 0 & 99 $\pm$ 0\\\hline
    
  \end{tabular}
  \caption{Percentage Accuracies for End-to-End, MC Models on Unicode Images.}
\end{table}

\clearpage
\section{Unicode Images}

We include all 100 grayscale unicode images that we used to construct the visual problem solving task. 

\begin{figure}[h!]
\centering
\includegraphics[width=\textwidth,height=\textheight,keepaspectratio]{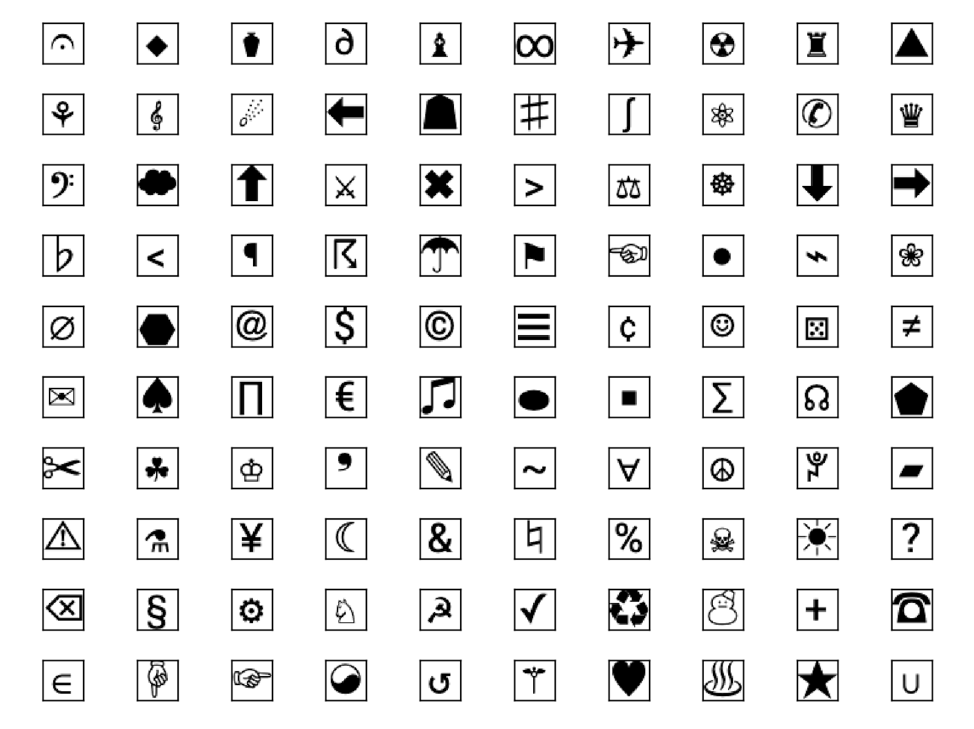}
\caption{The Full Set of 100 Grayscale Unicode Images}
\label{figure7}
\end{figure}

\clearpage
\section{Additional Examples of Generated Images}

The following figures include examples of predicted images (left), their nearest neighbors among the $N =100$ images according to Mean Squared Error (middle), and the target images (right).

\begin{figure}[h!]
\centering
\includegraphics[width=0.7\linewidth,height=0.7\textheight, keepaspectratio]{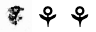}
\caption{Example of an accurate, yet imprecise, prediction.}
\label{figure8}
\end{figure}

\noindent In some cases in which a model's prediction was correct, the generated image was not very crisp or precise, though not indecipherable from the image it was attempting to predict.

\begin{figure}[h!]
\centering
\includegraphics[width=0.7\linewidth,height=0.7\textheight, keepaspectratio]{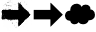}
\caption{Example of an inaccurate, yet precise, prediction.}
\label{figure9}
\end{figure}

\noindent In some cases in which a model's prediction was incorrect, the generated image was precise. This indicates that the model had strong certainty in its predictions, but involved filler-specific representations to solve the task. 

\begin{figure}[h!]
\centering
\includegraphics[width=0.7\linewidth,height=0.7\textheight, keepaspectratio]{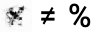}
\caption{Example of an incorrect, imprecise prediction.}
\label{figure10}
\end{figure}

\noindent In other cases in which a model's prediction was incorrect, it was not very clear what it was trying to predict.

\end{appendices}

\end{document}